\documentclass[letterpaper]{article}
\usepackage{aaai}
\usepackage{times}
\usepackage{helvet}
\usepackage{courier}
\usepackage{epic}
\usepackage{ecltree}
\usepackage{graphicx}
\usepackage{latexsym}
\usepackage{verbatim}

\begin{document}

\newlength{\halftextwidth}
\setlength{\halftextwidth}{0.47\textwidth}
\def\halffigsize{2.2in}
%\newlength{\thirdtextwidth}
%\setlength{\thirdtextwidth}{0.31\textwidth}
\def\thirdfigsize{1.5in}
\def\negvspace{0in}
\def\posvspace{0em}

\input epsf

%\usepackage{amssymb}

%\addtolength{\textwidth}{1.5in}
%\addtolength{\textwidth}{0.5in}
%\addtolength{\oddsidemargin}{-0.75in}
%\addtolength{\topmargin}{-0.5in}
%\addtolength{\textheight}{0.5in}

%\newcommand{\figref}[1]{Figure \ref{#1}}
%\newcommand{\tblref}[1]{Table \ref{#1}}
%\newcommand{\secref}[1]{Section \ref{#1}}
%\newcommand{\eqref}[1]{(\ref{#1})}
%\newcommand{\mygtrsim}{\gtrsim}

%\newtheorem{theorem}{Theorem}
%\newtheorem{definition}{Definition}
%\newtheorem{example}{Example}
%\newtheorem{theorem1}{Theorem}
%\newcommand{\proof}{\noindent {\bf Proof:\ \ }}
%\newcommand{\qed}{\mbox{$\Box$}}
%\newcommand{\qed}{\mbox{QED.}}

\newcommand{\set}{\mathcal}
\newcommand{\myset}[1]{\ensuremath{\mathcal #1}}

\renewcommand{\theenumii}{\alph{enumii}}
\renewcommand{\theenumiii}{\roman{enumiii}}
\newcommand{\figref}[1]{Figure \ref{#1}}
\newcommand{\tref}[1]{Table \ref{#1}}
\newcommand{\myOmit}[1]{}
\newcommand{\And}{\wedge}
\newcommand{\myldots}{.}

\newtheorem{mydefinition}{Definition}
\newtheorem{mytheorem}{Theorem}
\newtheorem{mytheorem1}{Theorem}
\newcommand{\myproof}{\noindent {\bf Proof:\ \ }}
\newcommand{\myqed}{\mbox{$\Box$}}

\newcommand{\mymod}{\mbox{\rm mod}}
\newcommand{\range}{\mbox{\sc Range}}
\newcommand{\roots}{\mbox{\sc Roots}}
\newcommand{\myiff}{\mbox{\rm iff}}
\newcommand{\alldifferent}{\mbox{\sc AllDifferent}}
\newcommand{\alldiff}{\mbox{\sc AllDifferent}}
\newcommand{\interdistance}{\mbox{\sc InterDistance}}
\newcommand{\permutation}{\mbox{\sc Permutation}}
\newcommand{\disjoint}{\mbox{\sc Disjoint}}
\newcommand{\cardpath}{\mbox{\sc CardPath}}
\newcommand{\CARDPATH}{\mbox{\sc CardPath}}
\newcommand{\knapsack}{\mbox{\sc Knapsack}}
\newcommand{\common}{\mbox{\sc Common}}
\newcommand{\uses}{\mbox{\sc Uses}}
\newcommand{\lex}{\mbox{\sc Lex}}
\newcommand{\usedby}{\mbox{\sc UsedBy}}
\newcommand{\nvalue}{\mbox{\sc NValue}}
\newcommand{\slide}{\mbox{\sc Slide}}
\newcommand{\SLIDE}{\mbox{\sc Slide}}
\newcommand{\circularslide}{\mbox{\sc Slide}_{\rm O}}
\newcommand{\among}{\mbox{\sc Among}}
\newcommand{\mysum}{\mbox{\sc Sum}}
\newcommand{\amongseq}{\mbox{\sc AmongSeq}}
\newcommand{\atmost}{\mbox{\sc AtMost}}
\newcommand{\atleast}{\mbox{\sc AtLeast}}
\newcommand{\element}{\mbox{\sc Element}}
\newcommand{\gcc}{\mbox{\sc Gcc}}
\newcommand{\egcc}{\mbox{\sc EGcc}}
\newcommand{\gsc}{\mbox{\sc Gsc}}
\newcommand{\contiguity}{\mbox{\sc Contiguity}}
\newcommand{\PRECEDENCE}{\mbox{\sc Precedence}}
\newcommand{\assignnvalues}{\mbox{\sc Assign\&NValues}}
\newcommand{\linksettobooleans}{\mbox{\sc LinkSet2Booleans}}
\newcommand{\domain}{\mbox{\sc Domain}}
\newcommand{\symalldiff}{\mbox{\sc SymAllDiff}}
\newcommand{\valsymbreak}{\mbox{\sc ValSymBreak}}

\newcommand{\slidingsum}{\mbox{\sc SlidingSum}}
\newcommand{\MaxIndex}{\mbox{\sc MaxIndex}}
\newcommand{\REGULAR}{\mbox{\sc Regular}}
\newcommand{\regular}{\mbox{\sc Regular}}
\newcommand{\STRETCH}{\mbox{\sc Stretch}}
\newcommand{\SLIDEOR}{\mbox{\sc SlideOr}}
\newcommand{\NAE}{\mbox{\sc NotAllEqual}}
\newcommand{\mymax}{\mbox{\rm max}}

\newcommand{\todo}[1]{{\tt (... #1 ...)}}

\title{The Parameterized Complexity of Global Constraints\thanks{Brahim Hnich
is supported by the Scientific and Technological Research Council of Turkey (TUBITAK) under Grant No. SOBAG-108K027.
Toby Walsh is funded by
the Australian Government's  Department of Broadband, Communications and the Digital Economy
and the 
Australian Research Council.}}
\author{C. Bessiere\\
LIRMM \\
Montpellier,\\ France \\
bessiere@lirmm.fr \And
E. Hebrard \\
4C\\
UCC, Ireland\\
ehebrard@4c.ucc.ie \And
B. Hnich\\
Izmir Uni. of Economics,\\
Izmir,\\ Turkey\\
brahim.hnich@ieu.edu.tr\And
Z. Kiziltan\\
CS Department \\
Uni. of Bologna, Italy \\
zeynep@cs.unibo.it \And
C.-G. Quimper\\
\'{E}cole Polytechnique\\
de Montr\'{e}al,\\ Canada\\
cquimper@alumni.uwaterloo.ca \And
T. Walsh\\
NICTA and UNSW\\
Sydney, Australia\\
toby.walsh@nicta.com.au}

\maketitle
\begin{abstract}
We argue that parameterized complexity is a useful
tool with which to study global constraints.
In particular, we show that many global
constraints %like \nvalue, \disjoint, and \common,
which are intractable to propagate completely have
natural parameters which make them fixed-parameter
tractable and which are easy to compute. This tractability tends either to be
the result of a simple dynamic program or
of a decomposition which has a strong backdoor of
bounded size. This strong backdoor is often a cycle cutset.
We also show that parameterized complexity
can be used to study other aspects of constraint programming
like symmetry breaking. For instance, we prove
that value symmetry is fixed-parameter
tractable to break in the number of symmetries. Finally, we
argue that parameterized complexity can be used to
derive results about the approximability of constraint
propagation.
\end{abstract}

%\vspace{-0.5em}
\section{Introduction}

One of the jewels of constraint programming is the notion of a global
constraint
(see, for example,
\cite{regin1,regin2,Bessiere-Regin97,regin4,beldiceanu2,fhkmwcp2002}).
Global constraints specify patterns that occur in many
real-world problems, and
come with efficient and effective propagation
algorithms for pruning the search space.
For instance, we often have decision variables which must
take different values (e.g. activities %in a scheduling problem
requiring the same resource must all be assigned different
times). Most constraint solvers therefore provide a global
\alldifferent\ constraint which is propagated efficiently
and effectively \cite{regin1}.
Unfortunately several common and useful global
constraints proposed in the past
have turned out to be intractable to propagate
completely (e.g.
\cite{quimper1,bhhwaaai2004,bhhkwijcai2005,bhhkwconstraint2006,szeider2,interdistance2}).
In this paper, we argue that we can understand
more about the source of this intractability by using
tools from parameterized complexity. The insights
gained from this analysis may lead to better search
methods as well as new %and efficient 
propagation algorithms.

\section{Formal background}

A constraint satisfaction problem (CSP) consists of a set of
variables, each with a finite domain of values, and a set of
constraints specifying allowed combinations of values for some
subset of variables. We presume any constraint can be checked in
polynomial time. We use capital letters for variables (e.g.
$X$, $Y$)%and $S$)
, and lower case for values (e.g. $d$ and $d_i$). We consider both
finite domain and set variables. A variable $X$ takes one value
from a finite domain of possible values $dom(X)$. A set variable
$S$ takes a set of values from a domain of possible sets. We 
view set variables as a vector of 0/1 finite domain variables
representing the characteristic function of the set. This
representation is equivalent to that of maintaining upper bound
($ub(S)$) and lower bound ($lb(S)$) on the potential and definite
elements in the set. However, it permits us to simplify our
analysis. % to just finite domain variables.

Constraint solvers typically use
backtracking search to explore the space
of partial assignments. After each assignment,
constraint propagation algorithms prune the search
space by enforcing local consistency properties like domain or bound
consistency. A constraint is \emph{domain
consistent} (\emph{DC})
iff when a variable is assigned any of the values in its domain, there
exist compatible values in the domains of all the other variables of
the constraint. Such values are called
a \emph{support}. A CSP is domain consistent iff every
constraint is domain consistent.
A constraint is \emph{bound consistent} (\emph{BC})
iff when a variable is
assigned the minimum (or maximum) value in its domain, there exist compatible
values between the minimum and maximum domain value
for all the other variables.
Such values are called
a \emph{bound support}. A CSP is bound consistent iff every
constraint is bound consistent.
%% A CSP is \emph{singleton domain consistent}
%% iff when a variable is assigned any of the values in its domain,
%% the resulting problem can be made domain consistent.
%When a constraint is binary (i.e. only over two variables),
%generalised arc consistency is simply called
%arc consistency (AC).
A \emph{global constraint} is one in which the number of variables
is not fixed. %a parameter.
%christian: we use this property in all our proofs
%A number of common and useful global
%constraints have been proposed.
For instance,
the $\nvalue([X_1,\ldots, X_n],N)$ constraint ensures that
the $n$ variables, $X_1$ to $X_n$,
take $N$ different values \cite{pachet1}.
%The number of variables, $n$ is a parameter.
The $\alldiff$ constraint \cite{regin1} is a special case of the
$\nvalue$ constraint in which  $N=n$.
%\cite{bhhwconstraint2007} have shown that
%enforcing domain consistency on a global constraint is NP-hard iff
%deciding the existence of a support is NP-complete.

\section{Parameterized complexity}

Recently, Bessiere {\it et al.} have
shown that a number of common global constraints
are intractable to propagate \cite{bhhkwconstraint2006}.
For instance, whilst enforcing
bound consistency on the \nvalue\ constraint is polynomial,
%enforcing 
domain consistency is NP-hard. 
We show here that the tools of parameterized complexity
can provide a more fine-grained
view of such complexity results.
These complexity tools help us to identify
more precisely what makes a global constraint
(in)tractable. The insights
gained %from such analyses 
may guide search -- for example,
we shall see that
they can identify small backdoors
on which to branch -- as well as
suggesting new propagation algorithms.

We introduce the necessary tools from
parameterized complexity theory.
A problem is \emph{fixed-parameter tractable}
(\emph{FPT}) if it can be solved in $O(f(k) n^c)$ time where
$f$ is \emph{any} computable function, $k$ is
some parameter, $c$ is a constant, and $n$ is
the size of the input.
For example, vertex cover (``Given a graph with $n$ vertices, is there
a subset of vertices of size $k$ or less that
cover each edge in the graph'') is NP-hard in general,
but fixed-parameter tractable with respect to $k$ since it
can be solved in $O({1.31951}^k k^2 + kn)$ time. %$ \cite{DowFelSte99}.
Hence, provided $k$ is small, vertex cover
can be solved effectively.

Downey {\it et al.} argue
\cite{DowFelSte99}
that about half the naturally parameterized
NP-hard problems in \cite{garey} are fixed-parameter
tractable including
3 out of the 6 basic problems.
Above $FPT$, Downey and Fellows have proposed
a hierarchy of fixed-parameter
\emph{intractable} problem classes:
$$ FPT \subseteq W[1] \subseteq W[2] \subseteq \ldots \subseteq XP $$
For instance, the clique problem
is $W[1]$-complete with respect to the size
of the clique, whilst the dominating set problem
is $W[2]$-complete with respect to the size
of the dominating set. $W[t]$ is characterized
by the maximum
number $t$ of unbounded fan-in gates on the input-output
path of a Boolean circuit specifying the problem.
There is considerable
evidence to suggest that $W[1]$-hardness implies parametric
intractability. In particular, the
halting problem for non-deterministic
Turing machines is $W[1]$-complete with
respect to the length of the accepting computation.

%{\bf - ??manu}
%We shall denote the parameterized version of a problem $A$
%by a parameter $k$ by $\langle A,k \rangle$.
%{\bf /??manu - }

\section{An example}

Parameterized complexity gives us
a more fine-grained view of the
complexity of propagating global
constraints.
Consider again the \nvalue\ constraint.
Whilst enforcing domain consistency
on the \nvalue\ constraint
is NP-hard in general, it is fixed-parameter tractable
to enforce in the total number of values in the domains.
In hindsight, it is perhaps obvious that propagating \nvalue\ is
easy when the number of values is fixed. However, many different
propagation algorithms for \nvalue\ have been proposed, 
and none of them prune all possible values
and are polynomial when the number of values is fixed.

\begin{mytheorem}
Enforcing domain consistency
on $\nvalue([X_1,\ldots,X_n],N)$
is fixed-parameter tractable in $k = |\bigcup_i dom(X_i)|$.
%
%{\bf - ??manu}
%$\langle \nvalue, |\bigcup_i dom(X_i)| \rangle$ is fixed-parameter tractable.
%{\bf - /??manu}
\end{mytheorem}
\myproof
We give an automaton for accepting
solutions to this constraint that scans
through $X_1$ to $X_n$ and then $N$. The states of this
automaton are all the possible sets of values that can be used by the
$X_i$,  plus one accepting state $F$. As there are $k$ possible values in the
domains of the $X_i$, there are $O(2^k)$ states.
The transition on seeing $X_i$ from state $q$ goes to state $q \cup \{X_i\}$.
Finally, we only accept a transition on seeing $N$ from state $q$ if
$|q|=N$. %, where $q_{n+1}$ is the state reached after reading $X_n$.
Quimper and Walsh have shown that we can efficiently 
enforce domain consistency wrt the assignments accepted
by such an automaton in $O(2^k nd)$ time where $d = \mymax \{ |dom(X_i)|\}$
using a simple decomposition \cite{qwcp07}.
\myqed

Thus, if the total number of values in the domains is small, 
\nvalue\ is tractable to propagate completely since propagation
takes polynomial time in the number of variables.
For such complexity results to be useful, the identified
parameter needs to be ``natural'' and potentially
small. 
It helps also if the parameter
is easy to compute. We can then build a
propagator which only decides to propagate
the constraint completely when it
is cheap to do so. As we shall see,
several global constraints are
fixed-parameter tractable with respect to
the total number of values in the domains.
This parameter is easy to compute, and is
often small in practice. 
For instance, many radio frequency link
assignment problems only require half a dozen values. 
We shall also see that dynamic programming is
often a means (as here) to show that a global
constraint is fixed-parameter tractable.

Other parameters can help identify features
that make a global constraint intractable.
For instance,
domain consistency on
the \nvalue\ constraint is intractable to
enforce when the fixed parameter is the maximum
number of values that can be used by the $X_i$. %being counted.

\begin{mytheorem}
\label{theorem::nvalueIntractable}
Enforcing domain consistency
on $\nvalue([X_1,\ldots,X_n],N)$
is $W[2]$-hard in $k = \mymax ( dom(N))$.
%
%{\bf - ??manu}
%$\langle \nvalue, \mymax(dom(N)) \rangle$ is $W[2]$-hard.
%{\bf - /??manu}
\end{mytheorem}
\myproof
Hitting set is $W[2]$-complete in the
size of the hitting set.
%Each support for the \nvalue\ constraint
%corresponds to a hitting set.
%{\bf - manu}
Given a collection of sets $\{S_1,\ldots,S_n\}$
and an integer $k$,
consider the set of variables $\{X_1,\ldots,X_n,N\}$
such that $dom(N) = \{0..k\}$ and $\forall i,~dom(X_i) = S_i$.
It is easy to see that a solution of
$\nvalue([X_1,\ldots,X_n],N)$ corresponds to
a hitting set of cardinality $k$ or less.
%{\bf /manu - }
Thus,
we can reduce hitting set to \nvalue\
immediately.
%{\bf I think we should mention what are the vertices, what are the
%edges, what is the parameter in the first problem, what is the
%parameter in the second parameter, and how is the parameter in the
%original problem function of the parameter in the transformed
%problem. Could someone remind me in which direction one needs to do
%the reduction? 8-) CG}
\myqed

Hence, we see that the complexity of
propagating the \nvalue\ constraint
comes from the potentially large number of
values in the domains of the $X_i$. %being counted.

\section{Backdoors}

As we shall see in the next
section, dynamic programming is a frequent way
to obtain fixed-parameter tractability results
for constraint propagators.
Another method %to obtain fixed-parameter tractability results
is to identify a decomposition of the global constraint
in which there is a strong backdoor containing
a bounded number of variables.
This backdoor is often a cycle cutset into
an acyclic (and thus polynomial)
subproblem. A \emph{strong backdoor} is
a subset of variables which give a polynomial
subproblem however they are instantiated \cite{backdoor}.
A \emph{cycle cutset} is a subset of variables which
break all cycles in the problem \cite{decpea87}.
Once the cycle cutset is instantiated, the problem
can be solved in a backtrack free manner using
a domain consistency algorithm.

Consider the global constraint, $\disjoint([X_1,\ldots,X_n],[Y_1,\ldots,Y_m])$
\cite{beldiceanu3}. This ensures that $X_i \neq Y_j$
for any $i$ and $j$. Such
a global constraint is useful in scheduling
and time-tabling problems. Enforcing domain consistency
on \disjoint\ is NP-hard \cite{bhhkwercim06}.
However, it is fixed-parameter tractable in
the total number of values in the domains.
To show this, we give a simple decomposition
which has a strong backdoor of bounded size that is
a cycle cutset. In fact, we give a slightly
stronger result. Enforcing domain
consistency
on \disjoint\ is fixed-parameter tractable in
the size of the intersection of the domains
of $X_i$ and $Y_j$. Clearly if the total
number of values is bounded then the intersection
is too.

\begin{mytheorem}
Enforcing domain consistency
on $\disjoint([X_1,\ldots,X_n],[Y_1,\ldots,Y_m])$
is fixed-parameter tractable
in $k = | \bigcup_i dom(X_i) \cap \bigcup_j dom(Y_j)|$.
%
%{\bf - ??manu}
%$\langle \disjoint, | \bigcup_i dom(X_i) \cap \bigcup_j dom(Y_j)|) \rangle$
%is fixed-parameter tractable.
%{\bf - /??manu}
\end{mytheorem}
\myproof Without loss of generality, we assume $n \geq m$.  Consider a
set variable $S$ and the decomposition: $X_i \in S$ and $Y_j \not\in
S$. Recall that a set variable %like $S$ 
can be viewed as a vector of
0/1 variables representing the characteristic function. Let $S_v=1$
iff $v \in S$.
%Then once $S_v$ is set for $v \in \bigcup_i dom(X_i)
%\cap \bigcup_j dom(Y_j)$,
%the cycles in the decomposition are broken,
%and domain consistency on the decomposition will check if the problem
%has support.
%Hence,
%once this cycle cutset is instantiated, each constraint of the
%decomposition becomes a unary constraint.
%*********christian
Then, the set $\{S_v~|~v \in \bigcup_i dom(X_i)\cap \bigcup_j
dom(Y_j)\}$ is a strong backdoor because once these
variables are set, the $X_i$ and $Y_j$ are disconnected and domain
consistency on the decomposition prunes the same values as on the
original constraint.
%********
Detecting the supports can
therefore be done in $O(nd)$ time
where $d= \mymax({|dom(X_i)|}_{i \in [1,n]} \cup {|dom(Y_j)|}_{j \in [1,m]})$.
Since there
are at most $O(2^k)$ possible instantiations for
the
%cycle cutset,
strong backdoor, enforcing domain consistency on
the \disjoint\ constraint can be achieved by calling $O(2^k)$ times domain
consistency on the decomposition and  taking the union of  the $2^k$
domain consistent domains  obtained. This
takes
$O(2^k nd)$ time.
\myqed

\section{Other examples}

We give some other examples of global
constraints which are fixed-parameter tractable
to propagate.

\subsection{Uses}

The global constraint $\uses([X_1,\ldots,X_n],[Y_1,\ldots,Y_m])$ holds
iff $\{ X_i \ | 1 \leq i \leq n \} \subseteq \{ Y_j \ | \ 1 \leq j \leq m\}$.
That is, the $X_i$ use only values used by the $Y_j$.
Enforcing domain consistency on such a constraint
is NP-hard \cite{bhhkwijcai2005}.
However, it is fixed-parameter tractable to propagate
in the total number of values in the domain of $Y_j$.
We let $k = | \bigcup_j dom(Y_j) |$.
We give an automaton for accepting
solutions to the \uses\ constraint that scans
through $Y_1$ to $Y_m$ and then $X_1$ to $X_n$. The states of this
automaton are all the possible sets of values that can be used
by $Y_j$. As there are $k$ possible values in the
domains of the $Y_j$, there are $O(2^k)$ possible states.
The transition on seeing $Y_j$  from state $q$ goes to $q \cup \{Y_j\}$.
We also only accept a transition on seeing $X_i$ from a state $q$
% to state $q$ itself  
if $X_i \in q$. This transition goes to state $q$ itself. 
%where $q_{m+1}$ is the  state reached after seeing $Y_m$.
We can therefore enforce domain consistency using the simple
decomposition in \cite{qwcp07}
in $O(2^k (n+m)d)$ time where $d = \mymax \{ |dom(X_i)|\}
\cup \{|dom(Y_j)|\}$.

\subsection{Among}

The \among\ constraint was introduced in CHIP to model resource
allocation problems like car sequencing \cite{beldiceanu2}. It
counts the number of variables using values from a given set.
$\among([X_1,\ldots,X_n],[d_1,\ldots,d_m],N)$ holds iff $N = |\{ i
\ | \ X_i=d_j, 1 \leq i \leq n, 1 \leq j \leq m\}|$. We here consider
 a generalisation of  \among\ where instead of the fixed values 
$d_j$ we have  a set variable $S$. 
That is, $\among([X_1,\ldots,X_n],S,N)$ holds iff $N = |\{ i
\ | \ X_i \in S, 1 \leq i \leq n\}|$. 
Enforcing domain consistency
on this extended version of \among\ is NP-hard
\cite{bhhkwercim06}. However, it is fixed-parameter tractable to
propagate in $k= |ub(S) \setminus lb(S)|$. \among\ can be decomposed into $(X_i\in
S)\leftrightarrow (B_i=1),\forall i$, and $\sum_i B_i=N$, where
$B_i$ are additional Boolean variables. 
(Note that the sum constraint is
polynomial to propagate when it sums   Boolean variables.)
$S$ is a cycle cutset of
the decomposition. Thus, once $S$ is set, domain consistency on
the decomposition is equivalent to domain consistency on the
original \among\ constraint. 
Since there are at most $O(2^k)$
possible instantiations for $S$, enforcing domain consistency on
the \among\ constraint can be achieved by calling $O(2^k)$ times
domain consistency on the decomposition and  making the union of
the $2^k$ domain consistent domains  obtained. This takes $O(2^k
nd)$ time.

\subsection{Roots}

Many counting and occurrence constraints can be specified using
the global \roots\ constraint \cite{bhhkwijcai2005}.
$\roots([X_1,\ldots,X_n],S,T)$ holds iff $S = \{ i \ | \ X_i \in
T\}$. As before, we consider $S$ and $T$ as shorthand for the
vector of 0/1 variables representing the associated characteristic
function. \roots\ can specify a wide range of other global
constraints including the \among, \atmost, \atleast,
\uses, \domain\ and \contiguity\ constraints.
Enforcing domain consistency on \roots\ is NP-hard
\cite{bhhkwcp2006}. However, it is fixed-parameter tractable to
propagate in $k= |ub(T) \setminus lb(T)|$. \roots\ can be
decomposed into $(i\in S)\leftrightarrow (X_i\in T),\forall i$.
$T$ is a cycle cutset of the decomposition. Thus, once $T$ is set,
domain consistency on the decomposition is equivalent to domain
consistency on the original \roots\ constraint. Since there are at
most $O(2^k)$ possible instantiations for $T$, enforcing domain
consistency on the \roots\ constraint can be achieved by calling
$O(2^k)$ times domain consistency on the decomposition and  making
the union of  the $2^k$ domain consistent domains  obtained. This
takes $O(2^k nd)$ time.

%
%
%\subsection{Sum}
%
%
%The constraint $\mysum(X_1,\ldots,X_n,Y)$ holds iff $Y=\sum_{i\in
%  1..n} X_i$. We assume that every variable $X_i$ is non-negative.
%Domain consistency on the $\mysum$ constraint is NP-hard (reduction of
%the {\sc SubsetSum} problem). However, \mysum\ is fixed-parameter
%tractable to propagate in $k=log_2(\max(dom(Y)))$.  We give an
%automaton inspired by~\cite{trick} for accepting solutions to this
%constraint that scans through $X_1$ to $X_n$ and then $Y$. We start
%with the state $q_0^0$. If $q_{i-1}^v$ is a state, $c \in dom(X_i)$,
%and $v + c \leq \max(dom(Y))$ then there is a state $q_i^{v+c}$ and
%the transition from state $q_{i-1}^v$ on reading $c$ goes to state
%$q_i^{v+c}$. There is a transition from state $q_n^c$ to the final
%state $F$ for every $c \in dom(Y)$. The state $q_0^0$ is the initial
%state and $F$ is the unique final state. Let $d$ be the cardinality of
%the largest domain among $dom(X_1), \ldots, dom(X_n)$ and
%$dom(Y)$. There are $O(2^kn)$ states in this automaton and $O(d)$
%transitions per state. Since the automaton is a layered graph with
%$n+1$ layers (one per variable), the propagator for the \REGULAR\
%constraint based on dynamic programming \cite{pesant1} can enforce
%domain consistency in $O(2^knd)$ time.

\section{Bound consistency}

Often bound consistency
on a global constraint is tractable but
domain consistency is NP-hard to enforce.
For instance, as we observed before,
bound consistency on the \nvalue\ constraint is polynomial,
but domain consistency is NP-hard \cite{bhhkwconstraint2006}.
As a second example,
bound consistency on the \interdistance\ constraint is polynomial,
but domain consistency is NP-hard \cite{interdistance,interdistance2}.
$\interdistance([X_1,\ldots,X_n],p)$ holds
iff $|X_i - X_j| \geq p$ for $i \neq j$. This global
constraint is useful in scheduling problems like
runway sequencing. As a third example,
the extended global cardinality
constraint, $\egcc([X_1,\ldots,X_n],[O_1,\ldots,O_m])$
ensures $O_j = |\{ i \ | \ X_i=j\}|$ for all $j$.
Enforcing bound consistency on the $O_j$ and domain
consistency on the $X_i$ is polynomial,
but enforcing domain consistency on all variables
is NP-hard \cite{quimper1}.
As a fourth and final example,
bound consistency on a linear equation with coefficients set to one is polynomial,
but domain consistency is NP-hard.

%We can give a general fixed-parameter tractability
%result for such global constraints.
%The parameter is simply the product of the number of intervals
%in the domains of the variables. This measures
%how close the problem is to bound consistency.
%If there are many ``holes'' in the domains,
%then we are far from a bound consistency problem
%and the problem is intractable.
%We define $intervals(S)= |\{ v \in S \ | v + 1 \not\in S\}|$.
%If $S$ contains no holes then $intervals(S)=1$.
%If $S$ contains $p$ holes, then $intervals(S)=p+1$.
%
%
%\begin{mytheorem}
%Suppose enforcing bound consistency
%on a global constraint over $X_1$ to $X_n$
%is polynomial. Then enforcing domain consistency
%is fixed-parameter tractable in $k = \prod_i^n intervals(dom(X_i))$.
%
%{\bf - ??manu}
%Then $\langle GAC, \prod_i^n intervals(dom(X_i))) \rangle$
%is fixed-parameter tractable.
%{\bf - /??manu}
%\end{mytheorem}
%\myproof We give a decomposition with a strong backdoor.
%Consider $X_i$. Suppose the $j$th interval in $dom(X_i)$ runs
%from $l_j$ to $u_j$ (that is, $l_j-1 \not\in dom(X_i)$,
%$u_j+1 \not\in dom(X_i)$, and $[l_j, u_j] \subseteq dom(X_i)$).
%We introduce a variable $Z_i$. When $Z_i=j$,
%$X_i$ will be restricted to the $j$th interval.
%To ensure this, we post $Z_i=j$ iff $l_j \leq X_i \leq u_j$.
%The $Z_i$ are a strong backdoor into a subproblem which
%bound consistency can solve. The number of possible
%instantiations of the backdoor is bounded by
%$\prod_i intervals(dom(X_i))$.
%\myqed

%christian: I have reformulated this bound versu gac relationship.
We can give a general fixed-parameter tractability
result for such global constraints.
The parameter is 
%simply the product of the number of intervals
%in the domains of the variables. 
the sum of the number of non-interval domains 
and of the maximum number of ``holes'' in a domain.
This measures
how close the domains are to intervals.
If there are many 
%``holes'' 
holes
in the domains,
then we are far from intervals
and the problem is intractable.
We define $intervals(S)= |\{ v \in S \ |\ v + 1 \not\in S\}|$.
If $S$ contains no holes then $intervals(S)=1$.
If $S$ contains $p$ holes, then $intervals(S)=p+1$.

\begin{mytheorem}
Suppose enforcing bound consistency
on a global constraint over $X_1$ to $X_n$
is polynomial. Then enforcing domain consistency
is fixed-parameter tractable in $k = p + q$
where $p=\mymax(intervals(dom(X_i))$ and $q$ is 
the number of non-interval variables.
%Suppose enforcing bound consistency
%on a global constraint over $X_1$ to $X_n$
%is polynomial. Then enforcing domain consistency
%is fixed-parameter tractable in $k = \prod_i^n intervals(dom(X_i))$.
%%
%%{\bf - ??manu}
%%Then $\langle GAC, \prod_i^n intervals(dom(X_i))) \rangle$
%%is fixed-parameter tractable.
%%{\bf - /??manu}
\end{mytheorem}
\myproof We give a decomposition with a strong backdoor.
Consider $X_i$. Suppose the $j$th interval in $dom(X_i)$ runs
from $l_j$ to $u_j$ (that is, $l_j-1 \not\in dom(X_i)$,
$u_j+1 \not\in dom(X_i)$, and $[l_j, u_j] \subseteq dom(X_i)$).
We introduce a variable $Z_i$. When $Z_i=j$,
$X_i$ will be restricted to the $j$th interval.
To ensure this, we post $(Z_i=j)\leftrightarrow (l_j \leq X_i \leq u_j)$.
By definition of bound consistency,  when domains are all intervals,
a  constraint is bound consistent iff it contains at
least a satisfying tuple.
Thus, the  $Z_i$ are a strong backdoor into a subproblem which
bound consistency can solve because when all $Z_i$ are set, all $X_i$
become intervals.
Checking if a value $v$ for $X_i$ is domain  consistent on the global
constraint is done by instantiating $X_i$ to $v$ (which is an
interval), and by trying all the possible
instantiations of the backdoor, except $Z_i$, until bound consistency does not
fail. This means that a support contains $X_i=v$.
The total cost is   bounded by
$nd\cdot\prod_i intervals(dom(X_i))$ times the cost of bound
consistency on this constraint, with $\prod_i intervals(dom(X_i)) \leq p^q$.
\myqed

\section{Meta-constraints}

Parameterized complexity also provides
insight into propagators for {meta-constraints}.
A \emph{meta-constraint} is a constraint
that applies other constraints. For instance,
given a constraint $C$ of arity $p$, the meta-constraint
$\cardpath(N,[X_1,\ldots,X_n],C)$
holds iff $N$ of the constraints,
$C(X_1,\ldots,X_p)$ to
$C(X_{n-p+1},\ldots,X_n)$ hold
\cite{cardinality-path}.
This permits
us to specify, say, that we want at least
2 days ``off'' in every 7 along a sequence
of shifts. \cardpath\ can
encode a range of Boolean connectives since $N\geq1$ gives
disjunction, $N=1$ gives exclusive or, and $N=0$ gives negation.
It therefore has numerous applications in domains
including car-sequencing and rostering.

Enforcing domain consistency on
$\cardpath$ is NP-hard even when
enforcing domain
consistency on each $C$ is polynomial
\cite{bhhwconstraint2007}.
However, domain consistency is fixed-parameter
tractable 
to enforce 
%when we bound 
with respect to
the sum of the
arity of $C$ and the maximum domain size.

\begin{mytheorem}
Enforcing domain consistency on $\cardpath(N,[X_1,\ldots,X_n],C)$
is fixed-parameter tractable in $k = p + d$
where $p$ is the arity of $C$ and
$d=\mymax |\{dom(X_i)\}|$.
%
%{\bf - ??manu}
%Let $p$ be the arity of $C$ and
%$d$ the maximum domain size of a variable,
%$\langle \cardpath, p+d \rangle$
%is fixed-parameter tractable.
%{\bf - /??manu}
\end{mytheorem}
\myproof
%christian: I have added the counter to the automaton
We give an automaton for accepting solutions to this constraint that
scans through $X_1$ to $X_n$ and then read $N$. The states of this
automaton are the possible sequences of values of length smaller than
$p-1$ labelled by the value 0, plus $n+1$ copies of the sequences of
values of length $p-1$ labelled with the integers from 0 to $n$, and a
final accepting state $F$.
% Each copy of a sequence of length $p-1$ is labelled with a different integer from 0 to $n$.
The integer labels count the number of times $C$ has been satisfied so
far. For $l < p$, the transition on reading $v_l$ from the state
$[v_1, \ldots, v_{l-1}]$ (labelled 0) goes to the state $[v_1, \ldots,
v_{l-1}, v_{l}]$ with label 0. The transition on
reading $v_p$ from the state $[v_1,\ldots, v_{p-1}]$ with label $r$
goes to the state $[v_2, \ldots, v_p]$ with label $r'$, where $r'=r+1$
if $C(v_1, \ldots, v_p)$ is satisfied and $r'=r$ otherwise. Finally,
there is a transition reading character $r$ from any state labelled
with $r$ to the final state $F$. The state $F$ is the unique final
state and the initial state is the empty sequence $\epsilon$ (labelled
0). There are $O(nd^{p-1})$ distinct states. 
We can therefore enforce domain consistency using the
simple decomposition in \cite{qwcp07} in
$O(d^pn^2)$ time.
% Dynamic programming can be used to construct
% a support. The states of the dynamic
% program are the $k-1$ previous assignments.
% Given a $k$-tuple,
% $C$ is (by definition) polynomial to check.
% As there are $O(d^p)$ possible states,
% we can enforce domain consistency on \cardpath\
% in $O(d^p nd)$ time.
\myqed

This is another example of a fixed-parameter
tractability result 
%where we needed two parameters, $p$ and $d$ to be fixed.
%Suppose we just fix $d$. Enforcing
%domain consistency on \cardpath\ is now
%intractable.
with respect to two parameters, $p$ and $d$. 
However, \cardpath\ is not fixed-parameter tractable
with respect to just $d$. In fact it is NP-hard
when $d$ is fixed.

\begin{mytheorem}
Enforcing domain consistency on $\cardpath(N,[X_1,\ldots,X_n],C)$
is NP-hard even if $|\{dom(X_i)\}| \leq 2$
\end{mytheorem}
\myproof
The reduction used in the proof of NP-hardness
of \cardpath\ in Theorem 12 \cite{bhhwconstraint2007} uses
just 1 or 2 domain values for each variable.
\myqed

\section{Symmetry breaking}

Parameterized complexity also provides
insight into symmetry breaking.
Symmetry is a common feature of many
real-world problems that dramatically increases the size
of the search space if it is not
taken into consideration.
Consider, for instance, value symmetry.
A \emph{value symmetry} is a bijection $\sigma$ on
values that preserves solutions. That is, if $X_i = a_i$ for $1 \leq i \leq n$
is a solution then $X_i = \sigma(a_i)$ is also.
For example, if two values are interchangeable,
then any possible permutation of these values is a
symmetry.
A simple and effective
mechanism to deal with symmetry is to
add constraints to eliminate symmetric solutions
\cite{puget:Sym,clgrkr96}.
For example, given a set of value symmetries $\Sigma$,
we can eliminate all symmetric solutions
by posting the global constraint
$\valsymbreak(\Sigma,[X_1,\ldots,X_n])$. This ensures
that, for each $\sigma \in \Sigma$:
$$\langle X_1, \ldots, X_n \rangle \leq_{\rm lex}
\langle \sigma(X_1), \ldots, \sigma(X_n) \rangle $$
Enforcing domain consistency on such a global
symmetry breaking constraint is NP-hard \cite{wcp07}.
However, this complexity depends on the number
of symmetries. 
%If we can bound the number of symmetries,
%then breaking all value symmetry is tractable.
Breaking all value symmetry is fixed-parameter tractable
in the number of symmetries.

\begin{mytheorem}
Enforcing domain consistency
on $\valsymbreak(\Sigma,[X_1,\ldots,X_n])$
is fixed-parameter tractable
in $k = | \Sigma|$.
%
%{\bf - ??manu}
%Let $k$ be the number of symmetric values,
%$\langle \valsymbreak, k \rangle$
%is fixed-parameter tractable.
%{\bf - /??manu}
\end{mytheorem}
\myproof
We give an automaton for accepting
solutions to this constraint that scans
through $X_1$ to $X_n$. The states
of the automaton are the set of
value symmetries which have been broken
up to this point in the vector. For instance,
at the $i$th state, $\sigma$ is a symmetry
in the state iff $\langle X_1, \ldots, X_{i} \rangle <_{\rm lex}
\langle \sigma(X_1), \ldots, \sigma(X_i) \rangle $.
If we are in the state $q$, we accept $X_{i}$ if
$X_{i} \leq \sigma(X_{i})$ or ($X_i > \sigma(X_i)$ and $\sigma \in q$).
From the state $q$, on seeing $X_i$,
we move to $q \cup \{ \sigma \ | \ \sigma \in \Sigma,
X_i < \sigma(X_i) \}$.
There are $O(2^k)$ possible states.
We can therefore enforce domain consistency using the 
simple decomposition in \cite{qwcp07}
in $O(2^k nd)$ time where $d = \mymax \{ |dom(X_i)|\}$.
\myqed

\section{Approximate consistency}

Parameterized complexity also provides
insight into the approximability of constraint
propagation. For optimization problems, global constraints
can incorporate a variable taking the objective
value. We can use approximation algorithms
to filter partially such global constraints
\cite{approxconsistency}. For example,
consider the knapsack constraint, $\knapsack([X_1,\ldots,X_n],
[w_1,\ldots,w_n],C,
[p_1,\ldots,p_n],P)$ which holds iff:
\begin{eqnarray*}
\sum_{i=1}^n w_i X_i \leq C & \ {\rm and} \ &
\sum_{i=1}^n p_i X_i > P
\end{eqnarray*}
$C$ is the capacity and $P$ is the profit.
Based on a fully polynomial time
approximation scheme, Sellmann
gives a propagator based on dynamic programming
%for filtering the domains
which guarantees
an \emph{approximate consistency} which 
filters values which only have supports that
are a factor $\epsilon$ outside the optimal
profit. A \emph{fully polynomial time
approximation scheme} (\emph{FPTAS})
is an algorithm that computes an answer with
relative error $\epsilon$ in time polynomial
in the input length and in $1/\epsilon$.
A weaker notion is an \emph{efficient}
polynomial time
approximation scheme (\emph{efficient PTAS})
which is an algorithm that computes an answer with
relative error $\epsilon$ in time polynomial
in the input length and in some function of $\epsilon$.

We can use parameterized complexity results
to show that such approximate consistency
is intractable to achieve.  In particular,
we can exploit a theorem first proved by Bazgan
that if a problem has an efficient \emph{PTAS} then
it is in $FPT$ \cite{DowFelSte99}.
Consider
again the constraint $\nvalue([X_1,\ldots,X_n],N)$.
We might ask if we can approximately filter domains.
%the number of
%values, $N$?
%{\bf christian: i'm not sure this is what we need. If i understood, we
%  will prune values that don't have support close enough (epsilon) to
%  a value in dom(N) ?}

\begin{mytheorem}
There is no polynomial
algorithm for enforcing approximate consistency
on $\nvalue([X_1,\ldots,X_n],N)$
unless $FPT=W[2]$.
\end{mytheorem}
\myproof By Theorem~\ref{theorem::nvalueIntractable}, enforcing domain
consistency on $\nvalue([X_1,\ldots,X_n],N)$ is $W[2]$-hard in $k =
\mymax \{ dom(N)\}$.  By Bazgan's theorem, we cannot have an efficient
\emph{PTAS} (and thus \emph{FPTAS}) for this problem unless
$FPT=W[2]$.  \myqed

\section{Other related work}

In addition to the related work already mentioned,
there are a number of other related studies.
The analysis of (in)tractability has a long
history in constraint programming. Such
work has tended to focus on the structure
of the constraint graph (e.g.  \cite{freuder4,dechter7})
or on the semantics of the constraints (e.g. \cite{cooper2}).
However, these lines of research are concerned with a constraint
satisfaction problem as a whole, and do not say much about
global constraints.

For global constraints of bounded arity, asymptotic
analysis has been used to study
%the complexity of 
propagation both in general and
for constraints with a particular semantics. For example,
the GAC-Schema algorithm of \cite{Bessiere-Regin97}
has an $O(d^n)$ time complexity on constraints
of arity $n$ and domains of size $d$,
whilst the GAC algorithm of  \cite{regin1}
for the $n$-ary \alldiff\
constraint has $O(n^{\frac{3}{2}}d)$
time complexity.
For global constraints like the
{\sc Cumulative} and
{\sc Cycle} constraints, there are very immediate reductions from
bin packing and Hamiltonian circuit which demonstrate
that these constraints are intractable to propagate in general.
%It is therefore perhaps not surprising
%that there has been little comment in the past about
%their intractability. However, 
Bessiere {\it et al.}
showed that many other global constraints like \nvalue\ are
also intractable to propagate \cite{bhhwaaai2004}. More recently,
Samer and Szeider have studied the parameterized
complexity of the \egcc\ constraint \cite{szeider2}.
They show it is fixed-parameter tractable to enforce
domain consistency in the tree-width of the value
graph and the largest possible number of occurrences,
but is $W[1]$-hard in just the tree-width.
Note that tree-width itself is NP-hard to compute.

\section{Conclusions}

We have argued that parameterized complexity is a useful
tool with which to study global constraints.
In particular, we have shown that many global
constraints like \nvalue, \disjoint, and \roots,
which are intractable to propagate completely have
natural parameters which make them fixed-parameter
tractable. This tractability tends either to be
the result of a simple dynamic program or
of a decomposition which has a strong backdoor of
bounded size. This strong backdoor is often a cycle cutset.
We also showed that parameterized complexity
can be used to study other aspects of constraint programming like
symmetry breaking. For instance, we proved
that value symmetry is fixed-parameter
tractable to break in the number of symmetries. Finally, we
argued that parameterized complexity can be used to
derive results about the approximability of constraint
propagation. For example, we cannot enforce
an approximate consistency within a guaranteed factor
for the \nvalue\ constraint.

%There are many directions for future work.
The insights provided by this work can help design
new search methods. 
For example,  \nvalue, \disjoint, \uses, \among\ and 
\roots\ all have efficient propagators when the total number 
of value is small with respect to the number of variables.
We might therefore build a
propagator that propagates partially using
a decomposition 
if the total number of values is large,
and  calls a complete method otherwise.
We might also exploit their decompositions
by branching on the backdoor variables. 
%This is likely to be a good strategy 
%as it simulates the fixed-parameter tractable 
%algorithm. 
Finally, when we 
have an efficient bound consistency propagator,
it may be worthwhile ``eliminating'' holes in the
domains by branching on those variables whose
domains have holes, or by introducing variables
to represent the intervals %within a domain 
without holes
and branching on these introduced variables. 
In the longer term, we hope to apply
other ideas about tractability from parameterized complexity
like reduction to a problem kernel.

\bibliographystyle{aaai}
%\bibliography{/Users/twalsh/Documents/biblio/a-z,/Users/twalsh/Documents/biblio/pub}
%\bibliography{pub,a-z}

%%\bibliographystyle{alpha}
\bibliography{/home/tw/biblio/a-z,/home/tw/biblio/pub}
%\bibliography{biblio,/home/tw/biblio/a-z,/home/tw/biblio/pub}
%%\bibliography{/n/endjinn/u6/tw/biblio/a-z,/n/endjinn/u6/tw/biblio/pub}
%%\bibliography{/usr/tw/biblio/a-z,/usr/tw/biblio/pub}
%\bibliography{/Users/tw/Documents/biblio/a-z,/Users/tw/Documents/biblio/pub}
%\bibliography{/Users/twalsh/Documents/biblio/a-z,/Users/twalsh/Documents/biblio/pub}
%%\bibliography{/home/arp/disk1/tw/biblio/a-z,/home/arp/disk1/tw/biblio/pub}
%%\bibliography{/u6/tw/biblio/a-z,/u6/tw/biblio/pub}
%%\bibliography{/usr/local/users/tw/biblio/a-z,/usr/local/users/tw/biblio/pub}
%\bibliography{biblio}

%\end{thebibliography}

\end{document}